\pdfoutput=1

\documentclass[11pt]{article}

\usepackage[]{acl}

\usepackage{times}
\usepackage{latexsym}

\usepackage[T1]{fontenc}

\usepackage[utf8]{inputenc}

\usepackage{microtype}

\usepackage{amsmath,amsfonts,bm}
\usepackage{hyperref}
\usepackage{url}
\usepackage{subfigure}
\usepackage{graphicx}
\usepackage{multirow}
\usepackage{wrapfig}
\usepackage{booktabs}
\usepackage{enumitem}

\newcommand{\Modelsp}{KG-FiD }
\newcommand{\Model}{KG-FiD}

\title{\Model: Infusing Knowledge Graph in Fusion-in-Decoder for Open-Domain Question Answering}

\author{Donghan Yu$^{1}$\thanks{~~Work done during internship at Microsoft.} , Chenguang Zhu$^{2}$, Yuwei Fang$^{2}$, Wenhao Yu$^{3}$\footnotemark[1] , Shuohang Wang$^{2}$, \\  \textbf{Yichong Xu$^{2}$, Xiang Ren$^{4}$, Yiming Yang$^{1}$, Michael Zeng$^{2}$} \\
$^{1}$Carnegie Mellon University $^{2}$Microsoft Cognitive Services Research Group \\
$^{3}$University of Notre Dame  $^{4}$University of Southern California \\
$^{1}$\texttt{dyu2@cs.cmu.edu}, $^{2}$\texttt{chezhu@microsoft.com}
}

\begin{document}
\maketitle

\begin{abstract}

Current Open-Domain Question Answering (ODQA) models typically include a  retrieving module and a reading module, where the retriever selects potentially relevant passages from open-source documents for a given question,  and  the  reader  produces  an answer  based  on  the  retrieved  passages. The recently proposed Fusion-in-Decoder (FiD) framework is a representative example, which is built on top of a dense passage retriever and a generative reader, achieving the state-of-the-art performance. In this paper we further improve the FiD approach by introducing a knowledge-enhanced version, namely \Model. Our new model uses a knowledge graph to establish the structural relationship among the retrieved passages, and a graph neural network (GNN) to re-rank the passages and select only a top few for further processing. 
Our experiments on common ODQA benchmark datasets (Natural Questions and TriviaQA) demonstrate that \Modelsp can achieve comparable or better performance in answer prediction than FiD, with less than 40\% of the computation cost.

\end{abstract}

\section{Introduction}

Open-Domain Question Answering (ODQA) is the task of answering natural language questions in open domains. A successful ODQA model relies on effective acquisition of world knowledge. A popular line of work treats a large collection of open-domain documents (such as Wikipedia articles) as the knowledge source, and design a ODQA system that consists of 
a \textit{retrieving module} and a \textit{reading module}. The retriever pulls out a small set of potentially relevant passages from the open-source documents for a given question,  and the  reader  produces  an answer  based  on  the  retrieved  passages \citep{dpr,guu2020realm,fidkd}.  An earlier example of this kind is DrQA \citep{drqa}, which used an traditional search engine based on the bag of words (BoW) document representation with TF-IDF term weighting, and a neural reader for extracting candidate answers for each query based on the dense embedding of the retrieved passages.
With the successful development of Pre-trained Language Models (PLMs) in neural network research, %
dense embedding based passage retrieval (DPR) models ~\citep{dpr,qu2021rocketqa} have shown superior performance over BoW/TF-IDF based retrieval models due to utilization of contextualized word embedding in DPR, and    %
generative QA readers~\citep{rag,roberts2020much} usually outperform extraction based readers~\citep{devlin2018bert,guu2020realm} due to the capability of the former in capturing lexical variants with a richer flexibility.

The recently proposed Fusion-in-Decoder (FiD) model \citep{fid} is representative of those methods with a DPR retriever and a generative reader, achieving the state-of-the-art results on ODQA evaluation benchmarks. FiD also significantly improved the scalability of the system over previous generative methods by encoding the retrieved passages independently instead of encoding the  concatenation of all retrieved passages (which was typical in previous methods).  

Inspired by the success of FiD, this paper aims further improvements of the state of the art of ODQA in the paradigm with a DPR retriever and a generative reader. Specifically, we point out two potential weaknesses or limitations of FiD as the rooms for improvements, and we propose a novel solution namely KG-FiD to address these issues with FiD.  The two issues are: 

\textit{Issue 1. The independent assumption among passages is not justified. } Notice that both the DPR retriever and the generative reader of FiD perform independent encoding of the retrieved passages, which means that they cannot leverage the semantic relationship among passages for passage embedding and answer generation even if such relational knowledge is available.
But we know that rich semantic connections between passages often provide clues for better answering questions~\citep{graphretriever}.

\textit{Issue 2. Efficiency Bottleneck}. For each input question, the FiD generative reader receives about 100 passages from the DPR module, with a relatively high computational cost.  For example, the inference per question takes more than 6 trillion floating-point operations.
Simply reducing the number of retrieved passages sent to the reader will
not be a good solution as it will significantly decrease the model performance \citep{fid}. How to overcome such inefficient computation issue is a challenging question for the success of FiD in realistic ODQA settings.

We propose to address both of the above issues with FiD by leveraging an  existing knowledge graph (KG) to establish relational dependencies among retrieved passages, and employing Graph Neural Networks (GNNs) to re-rank and prune retrieved passages for each query. We name our new approach as \Model.

Specifically, \Modelsp employs a two-stage passage reranking by applying GNN to model structural and semantic information of passages. Both stages rerank the input passages and only a few top-reranked passages are fed into subsequent modules. The first stage reranks passages returned by the retriever, where we use the passage embeddings generated by DPR as the initial GNN node representation. This allows reranking a much larger set of initial candidate passages to enhance coverage of answers. The second stage performs joint passage reranking and answer generation, where the node embeddings are initialized by the embeddings of passage-question pairs output from the reader encoder. This stage operates on a smaller candidate set but aims for more accurate reranking and passage pruning.

To improve the efficiency, in the second-stage reranking, our GNN model adopts representations from the intermediate layer in the reader encoder instead of the final layer to initiate passage node embeddings. Then only a few top reranked passages will be passed into the higher layers of encoder and the decoder for answer generation, while other passages will not be further processed. This is coupled with a joint training of passage reranking and answer generation. As shown in Section \ref{exp:main}, these strategies significantly reduce the computation cost while still maintaining a good QA performance.

Our experiments on ODQA benchmark datasets Natural Questions and TriviaQA demonstrate that \Modelsp can achieve comparable or better performance in answer prediction than FiD, 
with only 40\% of the computation cost of FiD.

\section{Related Work}

\paragraph{ODQA with text corpus}
ODQA usually assumes that a large external knowledge source is accessible and can be leveraged to help answer prediction. For example, previous works \citep{drqa,dpr,fid} mainly use Wikipedia as knowledge source which contains millions of text passages. In this case, current ODQA models mainly contains a retriever to select related passages and a reader to generate the answer. Thus, the follow-up works mainly aim to: (1) Improve the retriever: from sparse retrieval based on TF-IDF or BM25 \citep{drqa,bertserini} to dense retrieval \citep{dpr} based on contextualized embeddings generated by pre-trained language models (PLMs). Moreover, some further improvement are also proposed such as better training strategy \citep{qu2021rocketqa}, reranking based on retrieved passages \citep{r3,rerank-bert,rider}, and knowledge distillation from reader to retriever \citep{fidkd}; (2) Improve the reader: changing from Recurrent Neural Network \citep{drqa} to PLMs such as extractive reader BERT \citep{dpr,iyer2021reconsider,guu2020realm} and generative reader BART and T5 \citep{fid,rag}. Besides, some works \citep{guu2020realm,rag,jointtopk} have shown that additional unsupervised pre-training on retrieval-related language modeling tasks can further improve ODQA performance. However, none of these methods modeled the relationships among different passages.

\paragraph{ODQA with knowledge graph}
Besides the unstructured text corpus, world knowledge also exists in knowledge graphs (KGs), which represent entities and relations in a structural way and have been used in a variety of NLP tasks \citep{dekcor,yu2020jaket,xu2021does}. Some works \citep{berant2013semantic,GRAFT,sun2019pullnet,xiong2019improving} restrict the answer to be entities in the knowledge graph, while our work focus on more general ODQA setting where the answer can be any words or phrases. Under this setting, some recent efforts have been made to leverage knowledge graphs for ODQA \citep{graphretriever,pathretriever,zhou2020knowledgeaid}. For example, UniK-QA \citep{unikqa} transforms KG triplets into text sentences and combine them into text corpus, which loses structure information of KG. 
Other works use KG to build relationship among passages similar to ours. KAQA \citep{zhou2020knowledgeaid} use passage graph to propagate passage retrieve scores and answer span scores. Graph-Retriever \citep{graphretriever} iteratively retrieve passages based on the relationship between passages, and also use passage graph to improve passage selection in an extractive reader. However, applying KG to improve the recent advanced FiD framework remains unstudied.

\section{Method}

\begin{figure*}[!tp]
    \centering
    \includegraphics[width=16cm]{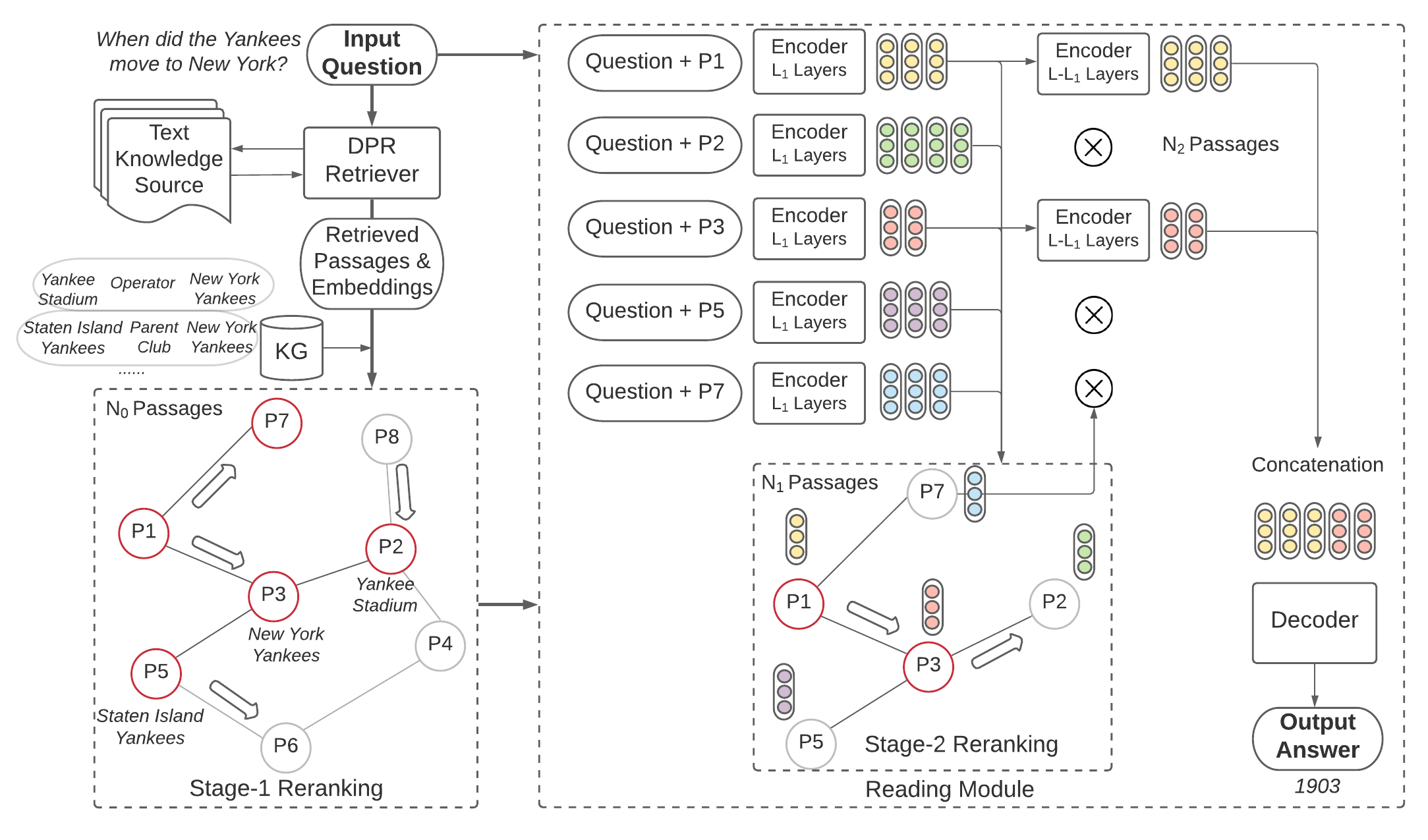}
    \caption{Overall Model Framework. P$i$ indicates the node of the passage originally ranked the $i$-th by the DPR retriever, with the article title below it. The left part shows passage retrieval by DPR, passage graph construction based on KG (Section \ref{sec:construct}) and stage-1 reranking (Section \ref{sec:rerank}). The right part shows joint stage-2 reranking and answer generation in the reading module (Section \ref{sec:effectiveness} and \ref{sec:efficiency}). }
    \label{fig:model}
\end{figure*}

In the following sections, we first introduce how to apply KG to build a graph structure among the retrieved passages (Section \ref{sec:construct}). Then we show how we adopt the graph-based stage-1 reranking with DPR retriever to improve passage retrieval (Section \ref{sec:rerank}). Next we introduce joint stage-2 reranking and answer generation in the reading module (Section \ref{sec:effectiveness}). Finally we illustrate the improvement of efficiency by using intermediate layer representation for stage-2 reranking~(Section \ref{sec:efficiency}). The overview of our framework is illustrated in Figure \ref{fig:model}.

\subsection{Construct Passage Graph using KG}
\label{sec:construct}
The intuition behind using KG is that there exists the structural relationship among the retrieved passages which can be captured by the KG.
Similar to~\cite{graphretriever}, we construct the passage graph where vertices are passages of text and the edges represent the relationships that are derived from the external KGs as $\mathcal{KG}=\{(e_h, r, e_t)\}$, where $e_h, r, e_t$ are the head entity, relation and tail entity of a triplet respectively.

First, we formalize the definition of a \textit{passage}. Following previous works \citep{wang2019multibert,dpr}, each article in the text corpus is split into multiple disjoint text blocks of 100 words called \textit{passages}, which serve as the basic retrieval units. 
We assume there is a one-one mapping between the KG entities and articles in the text corpus. Specifically, we use English Wikipedia as the text corpus and English Wikidata~\citep{vrandevcic2014wikidata} as the knowledge graph, since there exists an alignment between the two resources\footnote{Entity recognition and linking can be used if there is no such alignment.}. For example, for the article titled with ``New York Yankees'', it contains passages such as ``The New York Yankees are an American professional baseball team ...''. The article also corresponds to a KG entity with the same name as ``New York Yankees''. %

Then we define the mapping function $e = f(p)$, where the KG entity $e$ corresponds to the article which $p$ belongs to.
Note that one passage can only be mapped to one entity, but multiple passages could be mapped to the same entity. The final passage graph is defined as $\mathcal{G}=\{(p_i, p_j)\}$, where passages $p_i$ and $p_j$ are connected if and only if their mapped entities are directly connected in the KG, i.e., $(f(p_i), r, f(p_j))\in\mathcal{KG}$.

Since the total number of passages is very large, e.g., more than 20M in Wikipedia, constructing and maintaining a graph over all the passages is inefficient and memory-consuming. Thus, we build a passage graph on the fly for each question, based on the retrieved passages.

\subsection{Passage Retrieving \& Stage-1 Reranking}
\label{sec:rerank}

\textbf{DPR Retriever:} Our framework applies DPR \citep{dpr} as the retriever, which uses a BERT based passage encoder to encode all the $N$ passages in the text corpus $\{p_1, p_2,\cdots, p_N\}$. Suppose all the passage embeddings are fixed and stored in memory as $M \in \mathbb{R}^{N\times D}$ where $D$ is the hidden dimension:
\begin{align}
 M_i & = \text{BERT}(p_i) \text{ for } i \in \{1,2,\cdots N\}
 \label{eq:dpr}
\end{align}
For an input question $q$, DPR applies another BERT-based question encoder to obtain its representation $Q$, then it builds on FAISS~\citep{FAISS} to conduct fast dot-product similarity search between $Q$ and $M$, and returns $N_1$ ($N_1 \ll N$) passages with the highest similarity scores. 

\textbf{Stage-1 Reranking:} We see that the DPR retriever returns $N_1$ passages which are independently retrieved based on the similarity between the question and each passage, without considering inter-passage relationship.
Thus instead of directly retrieving $N_1$ passages for the reader, we propose to first retrieve $N_0$ ($N_0 > N_1$) passages, then rerank them and output top-$N_1$ reranked passages into the reader. 

Following Section \ref{sec:construct}, we construct a graph among the $N_0$ retrieved passages denoted as $\mathcal{G}_0$. We aim to rerank the retrieved passages based on both the structural information and the textual semantic information of them. 

To represent the semantic information of passages, one can use another pre-trained language model to encode the passage texts, but this will not only include lots of additional model parameters, but also incur heavy computational cost as $N_0$ can be large. To avoid both additional memory and computation cost, we propose to reuse the offline passage embeddings $M$ generated from the DPR retriever in Equation \ref{eq:dpr} as the initial node representation: $E_i^{(0)}=M_{r_i}$ where $\{r_i | i \in \{1,2,\cdots,N_0\}\}$ is the set of retrieved passage indices.

Then we employ a graph attention network (GAT) ~\citep{gat} with $L_g$ layers as GNN model to update representations for each node based on the passage graph and initial representation. 
The $l$-th layer of the GNN model updates the embedding of node $i$ as follows:
\begin{align}
    E_i^{(l)} = h(E_i^{(l-1)}, \{E_j^{(l-1)}\}_{(i,j) \in \mathcal{G}_0})
\label{eq:gat}
\end{align}
where $h$ is usually a non-linear learnable function which aggregates the embeddings of the node itself and its neighbor nodes.
The reranking score for each passage $p_{r_i}$ is calculated by $s_i^{\text{stage-1}} = Q^TE_i^{(L_g)}$, where $Q$ is the question embedding also generated by the DPR retriever. Then we sort the retrieved passages by the reranking scores, and input the top-$N_1$ passages into the reader. The training loss of passage ranking for each question is:
\begin{align}
    \mathcal{L}_{r}^{\text{stage-1}} = - \sum_{i=1}^{N_0} y_i \log \left( \frac{\exp(s_i^{\text{stage-1}})}{\sum_{j=1}^{N_0}\exp(s_j^{\text{stage-1}})} \right)
\label{eq:loss}
\end{align}
where $y_i=1$ if $p_{r_i}$ is the gold passage\footnote{We follow \citet{dpr} on the definition of gold passages.} that contains the answer, and 0 otherwise.

As we only add a lightweight graph neural network and reuse the pre-computed and static DPR passage embeddings, our reranking module can process a large number of candidate passages efficiently for each question. In experiments, we set $N_0=1000$ and $N_1=100$. %

\subsection{Joint Stage-2 Reranking and Answer Generation}
\label{sec:effectiveness}
In this section, we briefly introduce the vanilla FiD reading module before illustrating our joint reranking method. We suppose the reader takes $N_1$ retrieved passages $\{p_{a_1}, p_{a_2}, \cdots, p_{a_{N_1}}\}$ as input. %

\textbf{Vanilla FiD Reading Module:} We denote the hidden dimension as $H$ and number of encoder layers and decoder layers as $L$, FiD reader first separately encodes each passage $p_{a_i}$ concatenated with question $q$:
\begin{align}
    & \textbf{P}_i^{(0)} = \text{T5-Embed}(q + p_{a_i}) \in \mathbb{R}^{T_p \times H},  \\ 
    & \textbf{P}_i^{(l)} = \text{T5-Encoder}_l(\textbf{P}_i^{(l-1)}) \in \mathbb{R}^{T_p \times H}, 
\label{eq:fid-enc}
\end{align}
where $T_p$ is the sequence length of a passage concatenated with the question. $\text{T5-Embed}(\cdot)$ is the initial embedding layer of T5 model \citep{t5} and $\text{T5-Encoder}_l(\cdot)$ is the $l$-th layer of its encoder module. Then the token embeddings of all passages output from the last layer of the encoder are concatenated and sent to the decoder to generate the answer tokens $\textbf{A}$:
\begin{align}
    \textbf{A} = \text{T5-Decoder}\left([\textbf{P}_1^{(L)};\textbf{P}_2^{(L)};\cdots;\textbf{P}_{N_1}^{(L)}]\right)%
\label{eq:fid-dec}
\end{align}

\textbf{Stage-2 Reranking:} Note that vanilla FiD reader neglect the cross information among passages, and the joint modeling in the decoding process makes it vulnerable to the noisy irrelevant passages. Thus, we propose to leverage the passage graph to rerank the input $N_1$ passages during the encoding and only select top-$N_2$ ($N_2 < N_1$) reranked passages into the decoder, which is named as stage-2 reranking. 

Similar to stage-1 reranking, the reranking model is based on both the structural information and the textual semantic information of passages. We denote the passage graph as $\mathcal{G}_1$, which is a subgraph of $\mathcal{G}_0$. To avoid additional computation and memory cost, we propose to reuse the encoder-generated question-aware passage representation from FiD reader for passage reranking as it is already computed in Equation \ref{eq:fid-enc}. Specifically, the initial node embeddings $Z_i^{(0)}$ for passage $p_{a_i}$ comes from the first token embedding of the final layer in the FiD-Encoder, i.e., $Z_i^{(0)}=\textbf{P}_i^{(L)}(0) \in \mathbb{R}^{D}$. Then same as stage-1 reranking, we also employ a GAT ~\citep{gat} with $L_g$ layers as the graph neural network (GNN) model to update representations for each node based on the passage graph, similar to Equation \ref{eq:gat}: $Z^{(L_g)}=\text{GAT}(Z^{(0)}, \mathcal{G}_1^{\prime})$. The reranking score of passage $p_{a_i}$ is calculated by $s_i^{\text{stage-2}} = W^TZ_i^{(L_g)}$ where $W$ is a trainable model parameter. After reranking, only the final top-$N_2$ ($N_2 < N_1$) passages are sent for decoding. Suppose their indices are $\{g_1,g_2,\cdots,g_{N_2}\}$, the decoding process in Equation \ref{eq:fid-dec} becomes:
\begin{align}
    \textbf{A} = \text{T5-Decoder}\left([\textbf{P}_{g_1}^{(L)};\textbf{P}_{g_2}^{(L)};\cdots;\textbf{P}_{g_{N_2}}^{(L)}]\right)%
\label{eq:decode-2}
\end{align}
where \textbf{A} is the generated answer. Similar to stage-1 reranking, the training loss of passage ranking for each question is:
\begin{align}
    \mathcal{L}_{r}^{\text{stage-2}} = - \sum_{i=1}^{N_1} y_i \log \left( \frac{\exp(s_i^{\text{stage-2}})}{\sum_{j=1}^{N_1}\exp(s_j^{\text{stage-2}})} \right)
\label{eq:loss}
\end{align}
where $y_i=1$ if $p_{a_i}$ is the gold passage that contains the answer, and 0 otherwise. 

The passage reranking and answer generation are jointly trained. We denote the answer generation loss for each question is $\mathcal{L}_a$, then the final training loss of our reader module is $\mathcal{L} = \mathcal{L}_a + \lambda \mathcal{L}_r^{\text{stage-2}}$, where $\lambda$ is a hyper-parameter which controls the weight of reranking task in the total loss.

Note that the first stage reranking is based on DPR embeddings, which are are high-level (one vector per passage) and not further trained. While the second stage is based on reader-generated passage-question embeddings, which are semantic-level and trainable as part of the model output. Thus the second stage can better capture semantic information of passages and aims for more accurate reranking over a smaller candidate set. In the experiment, we set $N_1=100$ and $N_2=20$.

\subsection{Improving Efficiency via Intermediate Representation in Stage-2 Reranking}
\label{sec:efficiency}
Recall that in the stage-2 reranking, we take the passage representation from the last layer of  reader encoder for passage reranking. In this section, we propose to further reduce the computation cost by taking the intermediate layer representation rather than the last layer. The intuition is that answer generation task is more difficult than passage reranking which only needs to predict whether the passage contains the answer or not. Thus we may not need the representation from the whole encoder module for passage reranking.

Suppose we take the representation from the $L_1$-th layer ($1 \leq L_1 < L$), i.e., $Z_i^{(0)} = \textbf{P}_i^{(L_1)}(0)$ for $i \in \{1, 2, \cdots, N_1 \}$, and the reranking method remains the same. Then only the top $N_2$ ($N_2 < N_1$) reranked passages will go through the rest layers of FiD-encoder. Suppose their indices are $I_g = \{g_1,g_2,\cdots,g_{N_2}\}$, for $l \geq L_1+1$: 
\begin{align}
    \textbf{P}_i^{(l)} =
    \begin{cases}
    \text{T5-Encoder}_l(\textbf{P}_i^{(l-1)}) & \text{if } i \in I_g \\
    \text{Stop-Computing} & \text{else}
    \end{cases}
\end{align}
Then $\textbf{P}_{g_1}^{(L)},\textbf{P}_{g_2}^{(L)},\cdots,\textbf{P}_{g_{N_2}}^{(L)}$ are sent into the decoder for answer generation as in Equation \ref{eq:decode-2}.
In Section~\ref{exp:main}, we demonstrate this can reduce 60\% computation cost than the original FiD while keeping the on-par performance on two benchmark datasets.

\subsection{Analysis on Computational Complexity}
\label{sec:analysis}

Here we analyze the theoretical time complexity of our proposed \Modelsp compared to vanilla FiD. More practical computation cost comparison is shown in Appendix \ref{appendex:flops}. Because both the computations of DPR retrieving and stage-1 reranking are negligible compared to the reading part, we only analyze the reading module here. 

Suppose the length of answer sequence $\textbf{A}$ is denoted as $T_a$ and the average length of the passage (concatenated with question) is $T_p$. For vanilla FiD reader, the time complexity of the encoder module is $O(L \cdot N_1 \cdot T_p^2)$, where $L, N_1$ denote the number of encoder layers and the number of passages for reading. The square comes from the self-attention mechanism. The decoder time complexity is $O(L \cdot (N_1 \cdot T_p \cdot T_a + T_a^2))$, where $N_1 \cdot T_p \cdot T_a$ comes from the cross-attention mechanism. For our reading module, all the $N_1$ candidate passages are processed by the first $L_1$ layers of encoder. But only $N_2$ passages are processed by the remaining $L-L_1$ encoder layers and sent into the decoder. Thus, the encoder computation complexity becomes $O((L_1 \cdot N_1 + (L-L_1) \cdot N_2) \cdot T_p^2)$, and the decoder computation takes $O(L \cdot (N_2 \cdot T_p \cdot T_a + T_a^2))$. Because $L_1 < L, N_2 < N_1$, both the encoding and decoding of our method is more efficient than vanilla FiD.

Furthermore, the answer is usually much shorter than the passage (which is the case in our experiments), i.e., $T_a \ll T_p$. Then the decoding computation can be negligible compared to the encoding. In this case, the approximated ratio of saved computation cost brought by our proposed method is:
\begin{align*}
    S  & = 1 - \frac{(L_1 \cdot N_1+(L-L_1) \cdot N_2) \cdot T_p^2}{L \cdot N_1 \cdot T_p^2} \\ & =  (1-\frac{L_1}{L})(1-\frac{N_2}{N_1})
\end{align*}
This shows that we can reduce more computation cost by decreasing $L_1$ or $N_2$. For example, if setting $L_1=L/4, N_2=N_1/5$, we can reduce about $60\%$ of computation cost. More empirical results and discussions will be presented in Section \ref{exp:main}.

\section{Experiment}

In this section, we conduct extensive experiments on two most commonly-used ODQA benchmark datasets: Natural Questions (NQ)~\citep{naturequestions} which is based on Google Search Queries, and TriviaQA~\citep{joshi2017triviaqa} which contains questions from trivia and quiz-league websites. We follow the same setting as ~\cite{fid} to preprocess these datasets, which is introduced in Appendix \ref{appendix:dataset}. All our experiments are conducted on 8 Tesla A100 40GB GPUs. 

\subsection{Implementation Details}

\textbf{Knowledge Source:} Following \cite{dpr,fid}, we use the English Wikipedia as the text corpus, and apply the same preprocessing to divide them into disjoint passages with 100 words, which produces 21M passages in total. For the knowledge graph, we use English Wikidata. The number of aligned entities, relations and triplets among these entities are 2.7M, 974 and 14M respectively.

\textbf{Model Details:} For the retrieving module, we use the DPR retriever \citep{dpr} which contains two BERT (base) models for encoding question and passage separately.
For the GNN reranking models, we adopt 3-layer Graph Attention Networks (GAT) \citep{gat}. For the reading module, same as \cite{fid}, we initialize it with the pretrained T5-base and T5-large models \citep{t5}, and we name the former one as \Modelsp (base) and the latter one as \Modelsp (large). Our implementation is based on the HuggingFace Transformers library~\citep{wolf2019huggingface}. For number of passages, we set $N_0=1000,N_1=100,N_2=20$. The training process of our method is introduced in Appendix \ref{appendix:training}. More results about model design and hyper-parameter search is in Appendix \ref{appendix:exp}. 

\textbf{Evaluation:} We follow the standard evaluation metric of answer prediction in ODQA, which is the exact match score (EM)~\citep{rajpurkar2016squad}. A generated answer is considered correct if it matches any answer in the list of acceptable answers after normalization\footnote{The normalization includes lowercasing and removing articles, punctuation and
duplicated whitespace.}. For all the experiments, we conduct 5 runs with different random seeds and report the averaged scores.

\subsection{Baseline Methods}
 
We mainly compare \Modelsp with the baseline model FiD \citep{fid}. 
For other baselines, we compare with representative methods from each category: (1) not using external knowledge source: T5 \citep{roberts2020much} and GPT-3 \citep{gpt3}; (2) reranking-based methods: RIDER \citep{rider} and RECONSIDER \citep{iyer2021reconsider}; (3) leveraging knowledge graphs or graph information between passages: Graph-Retriever \citep{graphretriever}, Path-Retriever \citep{pathretriever}, KAQA \citep{zhou2020knowledgeaid}, and UniK-QA \citep{unikqa}. We also compare with methods (4) with additional large-scale pre-training: REALM \citep{guu2020realm}, RAG \citep{rag} and Joint Top-K \citep{jointtopk}.

\subsection{Main Results}
\label{exp:main}

\textbf{Comparison with Baselines:} \ Table \ref{tab:main} shows the results of our method and all baselines. We see that our proposed model \Modelsp consistently and significantly improves FiD on both NQ and TriviaQA datasets over both base and large model. Specifically, for large model, \Modelsp improves FiD by $1.5\%$ and $1.1\%$ on two datasets respectively, which has larger improvement compared to base model. We think the reason is that more expressive reader will also benefit the stage-2 reranking since the initial passage embeddings are generated by the reader encoder module. We also see that our proposed method outperforms all the baseline methods except UniK-QA \citep{unikqa}. However, UniK-QA uses additional knowledge source Wikipedia-Table for retrieval, which is highly related with the NQ dataset and makes it unfair to directly compare with our method.

\begin{table}[!tp]
\centering
\setlength{\tabcolsep}{3pt}
\begin{tabular}{lccc}
\hline
Model & \#params & NQ & TriviaQA \\  
\hline
T5 & 11B & 36.6 & - \\
GPT-3 (few-shot) & 175B & 29.9 & - \\
\hline
RIDER  & 626M & 48.3 & - \\
RECONSIDER  & 670M & 45.5 & 61.7 \\ \hline
Graph-Retriever%
& 110M & 34.7 & 55.8 \\ 
Path-Retriever%
& 445M & 31.7 & - \\
KAQA%
& 110M  & - & 66.6 \\  
UniK-QA%
$^{\star}$
& 990M & \textbf{54.0}$^{\star}$ & 64.1$^{\star}$ \\ 
\hline
REALM & 330M &  40.4 & -\\
RAG & 626M &  44.5 & 56.1 \\
Joint Top-K & 440M & 49.2 & 64.8\\ \hline
FiD (base)%
& 440M & 48.2 & 65.0 \\ 
FiD (large)%
& 990M & 51.4 & 67.6 \\ 
\hline
\multicolumn{4}{c}{Our Implementation} \\ \hline
FiD (base) & 440M & 48.8 & 66.2 \\ 
\Modelsp (base) & 443M & 49.6 & 66.7 \\
FiD (large) & 990M & 51.9 & 68.7 \\ 
\Modelsp (large) & 994M & \textbf{53.4} & \textbf{69.8} \\ \hline
\end{tabular}
\caption{Exact match score of different models over the test sets of NQ and TriviaQA datasets. $\star$ means that additional knowledge source Wikipedia-Tables is used in this method.
}
\label{tab:main}
\end{table}

\begin{table*}[!tp]
\centering
\begin{tabular}{lccccc}
\hline
\multirow{2}{*}{Model} & \multirow{2}{*}{\#FLOPs} & \multicolumn{2}{c}{NQ} & \multicolumn{2}{c}{TriviaQA} \\  \cmidrule(lr){3-4} \cmidrule(lr){5-6}   &        &  EM     &  Latency (s)                 & EM          &  Latency (s)           \\ \hline
FiD ($\text{N}_1$=40)    & 0.40x  & 50.3 &  0.74 (0.45x)  & 67.5 & 0.73 (0.44x) \\
  FiD ($\text{N}_1$=100)    & 1.00x  &  51.9 & 1.65 (1.00x)  & 68.7 & 1.66 (1.00x) \\ \hline
  \Modelsp ($\text{N}_1$=100, $\text{L}_1$=6)    & 0.38x & 52.0 & 0.70 (0.42x)  & 68.9 & 0.68 (0.41x)
         \\
 \Modelsp ($\text{N}_1$=100, $\text{L}_1$=12)    & 0.55x & 52.3 & 0.96 (0.58x) & 69.2 & 0.94 (0.57x) \\
 \Modelsp ($\text{N}_1$=100,  $\text{L}_1$=18)    & 0.72x  & 52.6 & 1.22 (0.74x) & 69.8 &  1.22 (0.73x) \\
  \Modelsp ($\text{N}_1$=100,  $\text{L}_1$=24)    & 0.90x  & 53.4 & 1.49 (0.90x)  & 69.8 & 1.48 (0.89x) \\ 
   \hline
\end{tabular}
\caption{Inference \#FLOPs, Latency (second) and Exact match score of FiD (large) and \Modelsp (large). $N_1$ is the number of passages into the reader and $L_1$ is the number of intermediate layers used for stage-2 reranking as introduced in Section \ref{sec:efficiency}. The details of flop computation is introduced in Appendix \ref{appendex:flops}.}
\label{tab:main2}
\end{table*}

\textbf{Efficiency \& Accuracy:} Table \ref{tab:main2} show the detailed comparison between our method and FiD in the large model version. The results of base model version is shown in Appendix \ref{appendix:exp}. Besides EM score, we also report the ratio of computation flops (\#FLOPs) and inference latency (per question). The detailed calculation of \#FLOPs is shown in Appendix \ref{appendex:flops}. From table \ref{tab:main2}, we see that (1) for \Model, decreasing $L_1$ can improve the computation efficiency as analyzed in Section \ref{sec:efficiency}, while increasing $L_1$ can improve the model performance. We think the performance improvement comes from the noise reduction of passage filtering. For a larger $L_1$, the passage embeddings for reranking will have a better quality so that the gold passages are less likely to be filtered out. (2) Simply reducing the number of passages $N_1$ into vanilla FiD reader can reduce computation cost, but the performance will also drop significantly (from 51.9 to 50.3 on NQ dataset). (3) Our model can achieve the performance on par with FiD with only $38\%$ of computation cost. When consuming the same amount of computations ($L_1=24$), our model significantly outperforms FiD on both NQ and TriviaQA datasets. These experiments demonstrate that our model is very flexible and can improve both the efficiency and effectiveness by changing $L_1$.

\subsection{Ablation Study}
\label{exp:ablation}

\textbf{Effect of Each Reranking Stage:} 
Since our proposed graph-based reranking method are applied in both retrieving stage (Section \ref{sec:rerank}) and reading stage (Section \ref{sec:effectiveness}). We conduct ablation study to validate the effectiveness of each one. Table \ref{tab:ablation} shows the experiment results by removing each module. We see the performance of \Modelsp drops when removing any of the two reranking modules, demonstrating both of them can improve model performance. Another thing we observe is that stage-1 reranking is more effective in base model while stage-2 reranking is more effective in large model. This is reasonable since stage-2 reranking relies on the effectiveness of reader encoder module, where the large model is usually better than the base model.

\begin{table}[!tp]
\centering
\begin{tabular}{lcccc}
\hline
\multirow{2}{*}{Model} & \multicolumn{2}{c}{NQ} & \multicolumn{2}{c}{TriviaQA} \\  \cmidrule(lr){2-3} \cmidrule(lr){4-5}          & base             & large                      & base           & large                     \\ \hline
  FiD     & 48.8 & 51.9 &   66.2 & 68.7 \\  \hline
  \Model     & 49.6 & 53.4 & 66.7 & 69.8 \\
   \ w/o Stage-1    & 49.3 & 53.1 & 66.2 & 69.5 \\
   \ w/o Stage-2    & 49.4  & 52.3 &  66.5 & 69.2 \\   \hline
\end{tabular}
\caption{Ablation study of our graph-based reranking method in two stages. EM scores are reported over NQ and Trivia datasets with both base and large model version.}
\label{tab:ablation}
\end{table}

\begin{figure}
    \centering
    \includegraphics[width=6.5cm]{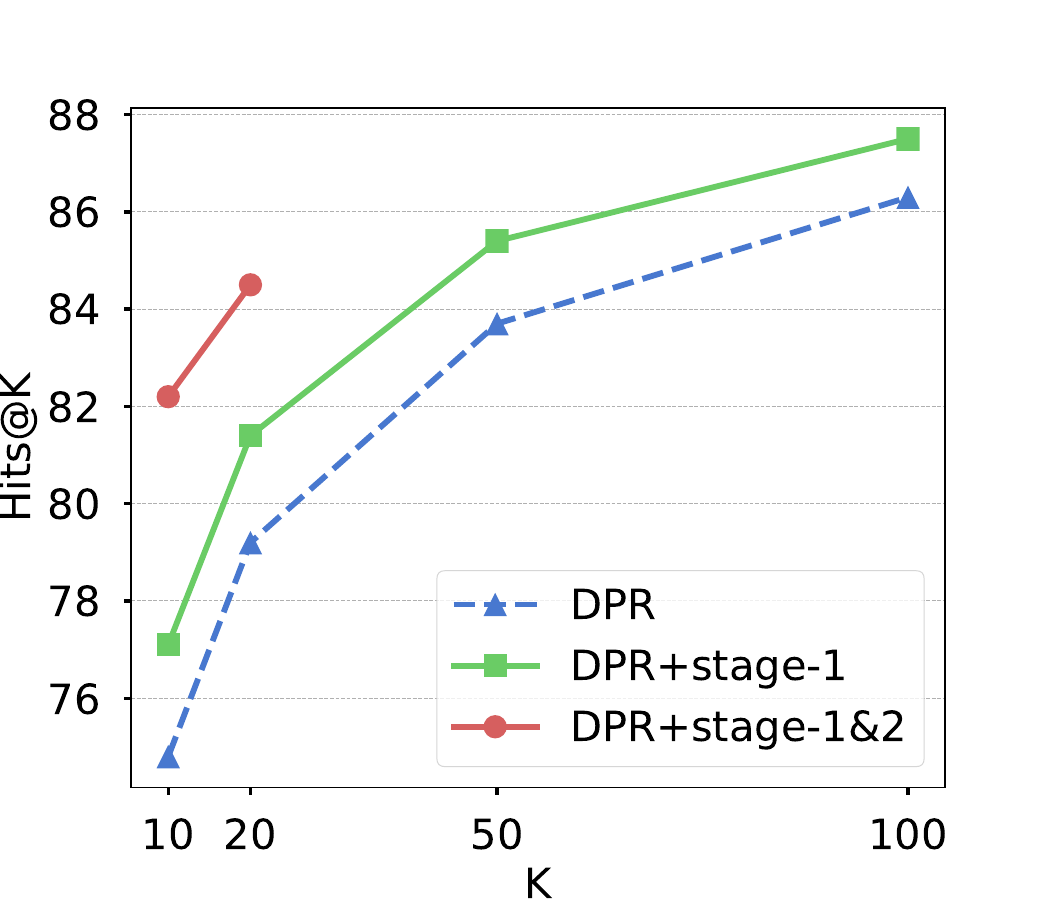}
    \caption{Passage ranking results over NQ test set of DPR retriever and our proposed two-stage rerankings over base model.}
    \label{fig:reranking}
\end{figure}

\textbf{Passage Ranking Results:} We additionally show that our proposed GNN reranking method can improve the passage retrieval results. This is demonstrated in Figure \ref{fig:reranking}, where we report Hits@K metric over NQ test set, measuring the percentage of top-K retrieved passages that contain the gold passages (passages that contain the answer). We see that DPR+stage-1 reranking consistently outperforms DPR for all the $K \in \{10, 20, 50, 100\}$. With two stages of reranking, the retrieval results are further improved for $K \in \{10, 20\}$ (We only cares about $K \leq 20$ for stage-2 reranking since $N_2=20$). This shows that such reranking can increase the rank of gold passages which are previously ranked lower by DPR retriever and improve the efficacy of passage pruning.

\section{Conclusion}
This work tackles the task of Open-Domain Question Answering. We focus on the current best performed framework FiD and propose a novel KG-based reranking method to enhance the cross-modeling between passages and improve computation efficiency. Our two-stage reranking methods reuses the passage representation generated by DPR retriver and the reader encoder and apply graph neural networks to compute reranking scores. We further propose to use the intermediate layer of encoder to reduce computation cost while still maintaining good performance. Experiments on Natural Questions and TriviaQA show that our model can significantly improve original FiD by $1.5\%$ exact match score and achieve on-par performance with FiD but reducing over $60\%$ of computation cost. 

\section{Acknowledgements}

We thank all the reviewers for their valuable comments. We also thank Woojeong Jin, Dong-Ho Lee, and Aaron Chan for useful discussions. Donghan Yu and Yiming Yang are supported in part by the United States Department of Energy via the Brookhaven National Laboratory under Contract No. 384608.

\bibliography{anthology,custom}
\bibliographystyle{acl_natbib}

\appendix

\section{Appendix}

\subsection{Dataset}
\label{appendix:dataset}

The datasets we use are Natural Questions (NQ) and TriviaQA. The open-domain version of NQ is obtained by discarding answers with more than 5 tokens. For TriviaQA, its \textit{unfiltered} version is used for ODQA. We also convert all letters of answers in lowercase except the first letter of each word on TriviaQA. When training on NQ, we sample the answer target among the given list of answers, while for TriviaQA, we use the unique
human-generated answer as generation target. For both datasets, we use the original validation data as
test data, and keep 10\% of the training set for validation.

\subsection{Preliminary Analysis}
\label{appendix:analysis}

\begin{figure*}[!tp]
    \centering
    \subfigure[Results on NQ]{
    \includegraphics[width=6.5cm]{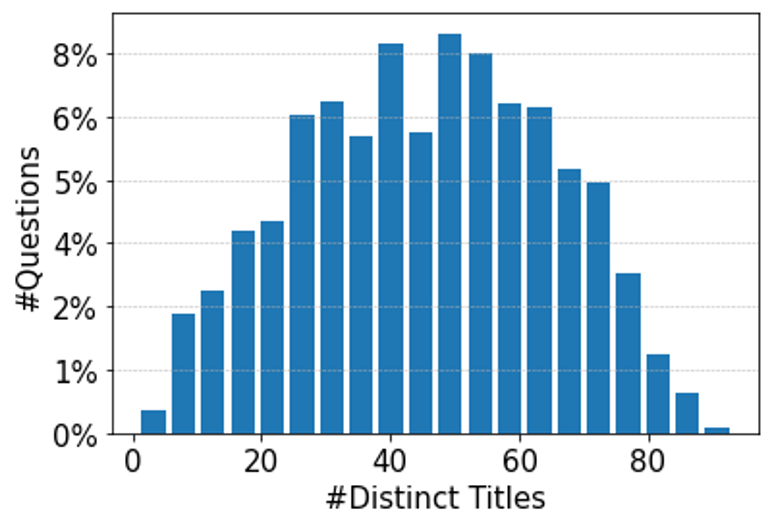}
    }
    \subfigure[Results on TriviaQA]{
    \includegraphics[width=6.6cm]{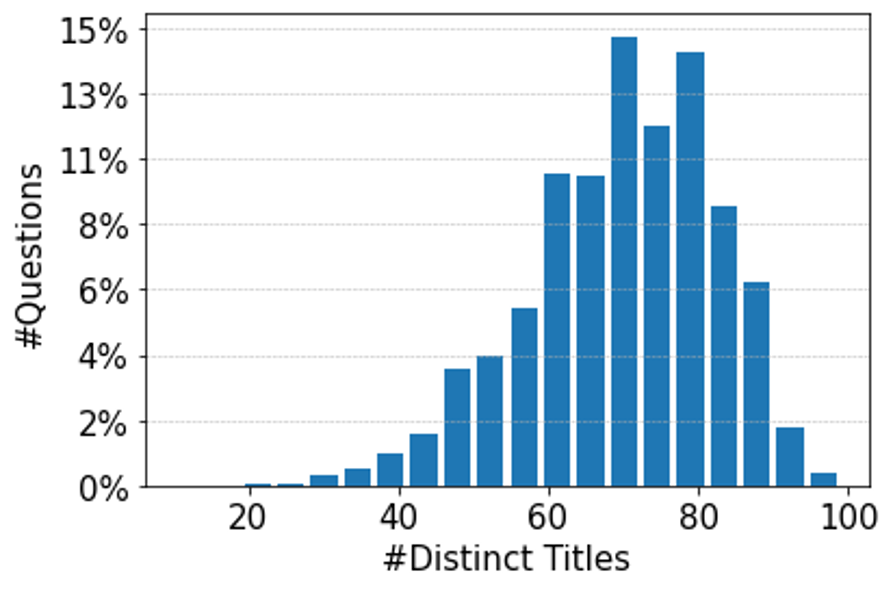}
    }\\
    \subfigure[Results on NQ]{
    \includegraphics[width=6.5cm]{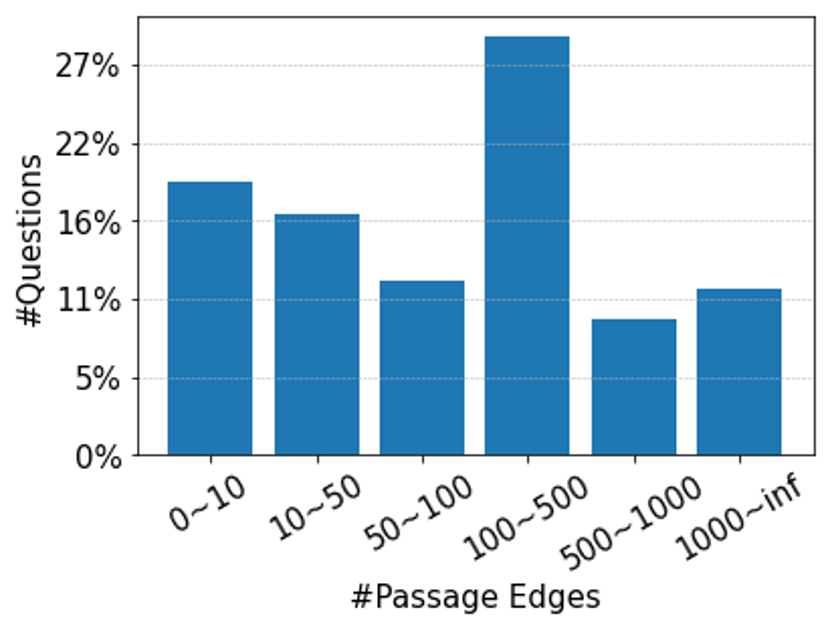}
    }
    \subfigure[Results on TriviaQA]{
    \includegraphics[width=6.6cm]{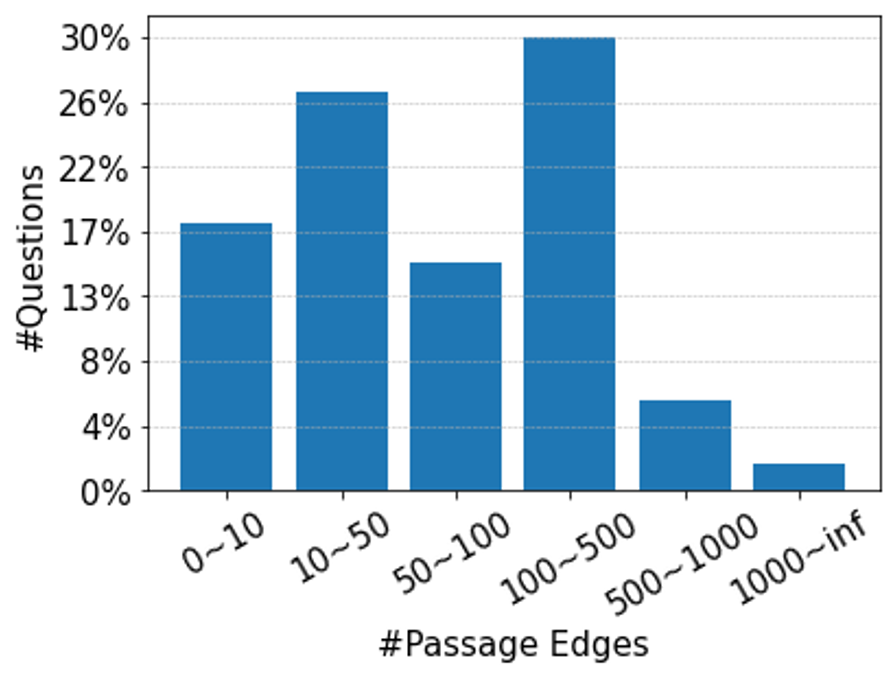}
    }
    \caption{Preliminary Analysis on the retrieved passages by DPR.}
    \label{fig:pre}
\end{figure*}

We conduct preliminary analysis on the graph constructed among passages. Note that for each question, we first apply the retriever to retrieve a few candidate passages, then build edge connection only among the retrieved passages, which means that the passage graph is question-specific.
Since the passage graph depends on the retrieved passages, before further utilizing the graph, we need avoid two trivia situations: (1) all the retrieved passages come from the same article; (2) The number of graph edges is very small. Thus we conduct statistics of the passage graphs on two ODQA benchmark datasets, which is shown in Figure \ref{fig:pre}. For each question, the number of retrieved passages is 100. We see that the two trivia situations only happen for a small portion of questions.

\subsection{Training Process}
\label{appendix:training}

For training our framework, we adopt the separate-training strategy to avoid out-of-memory issue: we first train the DPR model following its original paper, then freeze the DPR  model to train the stage-1 reranking module, and finally jointly train stage-2 reranking and reader part. For the training of stage-1 reranking, the optimizer is AdamW \citep{adamw} with learning rate as 1e-3 and linear-decay scheduler. The weight decay rate is 0.01. Batch size is set as 64. The number of total training steps is 15k, and the model is evaluated every 500 steps and the model with best validation results is saved as the final model. For the training of reading part, we adopt the same training setting except that the learning rate is 1e-4 for the base model and 5e-5 for the large model. We also adopt learning rate warm up with 1000 steps. 

\subsection{Additional Experiment Results}
\label{appendix:exp}

We show additional experiment results in this section, which includes the efficiency and performance comparison between FiD (base) and \Modelsp (base) shown in Table \ref{tab:main1}, and hyper-parameter search results listed below:

\begin{table*}[!tp]
\centering
\begin{tabular}{lccccc}
\hline
\multirow{2}{*}{Model} & \multirow{2}{*}{\#FLOPs} & \multicolumn{2}{c}{NQ} & \multicolumn{2}{c}{TriviaQA} \\  \cmidrule(lr){3-4} \cmidrule(lr){5-6}   &        & EM      &  Latency (s)                 & EM      &  Latency (s)           \\ \hline
FiD ($\text{N}_1$=40)    & 0.40x  & 47.2  & 0.27 (0.47x) & 64.1 & 0.27 (0.46x)  \\ 
  FiD ($\text{N}_1$=100)    & 1.00x  & 48.8 & 0.58 (1.00x) &  66.2 & 0.59 (1.00x) \\  \hline
  \Modelsp ($\text{N}_1$=100, $\text{L}_1$=3)    & 0.38x  & 48.4 & 0.27 (0.47x)  & 65.6 & 0.26 (0.44x) \\
   \Modelsp ($\text{N}_1$=100, $\text{L}_1$=6)    & 0.56x  & 49.0 & 0.35 (0.60x) & 66.1 & 0.34 (0.58x) \\
   \Modelsp ($\text{N}_1$=100, $\text{L}_1$=9)    & 0.73x  & 49.3 & 0.43 (0.74x)  & 66.3 &  0.43 (0.73x)\\
  \Modelsp ($\text{N}_1$=100, $\text{L}_1$=12)    & 0.91x  & 49.6 & 0.50 (0.86x) & 66.7 & 0.49 (0.83x) \\    \hline
\end{tabular}
\caption{Inference \#FLOPs, Latency (second) and Exact match score of FiD (base) and \Modelsp (base). $N_1$ is the number of passages into the reader and $L_1$ is the number of intermediate layers used for stage-2 reranking as introduced in Section \ref{sec:efficiency}. The details of flop computation is introduced in Appendix \ref{appendex:flops}.}
\label{tab:main1}
\end{table*}

\textbf{GNN Model Design:} We conduct tuning on the model type and number of layers of our GNN based reranking model. For efficiency, we rerank 100 passages returned by DPR retriever and search them based on the passage retrieval results. Table \ref{exp:search-2} shows the Hits scores for different choices. We see that GAT outperforms vanilla GCN model~\citep{gcn} which is reasonable since GAT leverage attention to reweight neighbor passages by their embeddings. The best choice for the number of GNN layers is 3. Note that other GNN models such as GIN~\citep{gin}, DGI~\citep{dgi} can also be applied here and we leave the further exploration of GNN models as future work.

\textbf{$N_2$ and $\lambda$}. For the stage-2 reranking part in Section \ref{sec:effectiveness}, we also conduct hyper-parameter search on the number of passages after filtering: $N_2 \in \{10, 20, 30\}$ and the weight of reranking loss when training the reading module: $\lambda \in \{0.01, 0.1, 1.0\}$. As shown in Table \ref{exp:search-1}, $N_2=20$ achieves better results than $N_2=10$, but further increasing $N_2$ does not bring performance gain while decreasing the efficiency of model since the number of passages to be processed by the decoder is increased. Thus we choose $N_2=20$. For the loss weight $\lambda$, we found that with its increment, the performance first increases then significantly drops. This shows that it's important to balance the weight of two training losses, as we want the model to learn better passage reranking while not overwhelming the training signal of answer generation.

\begin{table}[!tp]
\centering
\begin{tabular}{ccccc}
\hline
     Model & H@1 & H@5 & H@10 & H@20                         \\ \hline
  GCN & 49.1 & 69.7 & 75.7 & 79.9 \\
  GAT & 50.1 & 70.1 & 76.1 & 80.2
         \\ \hline
     \#Layers %
     \\ \hline
  1 & 49.0 & 69.7 & 75.8 & 79.8
 \\
  2 & 49.6 & 70.0 & 76.0 & 80.2 \\
  3  & 50.1 & 70.1 & 76.1 & 80.2 \\
  4 & 49.5 & 69.9 & 76.1 & 80.1 
         \\ \hline
\end{tabular}
\caption{Passage Retrieval Results on NQ dev data of our model under different GNN types and number of layers.}
\label{exp:search-2}
\end{table}

\begin{table}[!tp]
\centering
\begin{tabular}{lcccc}
\hline
     Model & $\text{N}_2$=10 & $\text{N}_2$=20 & $\text{N}_2$=30                    \\ \hline
\Model    & 47.6 & 48.0 & 48.0 \\ \hline
 
 & $\lambda$=0.01 & $\lambda$=0.1 & $\lambda$=1.0                    \\ \hline
\Model     & 47.7  & 48.0 & 46.6
         \\ \hline
\end{tabular}
\caption{EM scores on NQ dev data of our model under different choices of filtered passage numbers and weights of reranking loss.}
\label{exp:search-1}
\end{table}

\begin{table*}[!tp]
\centering
\begin{tabular}{lcccccc}
\hline
Model & Retrieving & \multicolumn{1}{c}{\begin{tabular}[c]{@{}c@{}}Stage-1 \\ Reranking\end{tabular}} & \multicolumn{1}{c}{\begin{tabular}[c]{@{}c@{}}Reader \\ Encoding\end{tabular}} & \multicolumn{1}{c}{\begin{tabular}[c]{@{}c@{}}Stage-2 \\ Reranking\end{tabular}} & \multicolumn{1}{c}{\begin{tabular}[c]{@{}c@{}}Reader \\ Decoding\end{tabular}} & All        \\ \hline
FiD    & 4.4 & - & 5772.3 &  - & 714.2 & 6491.0 (1.00x)  \\ \hline
  \Modelsp ($\text{L}_1$=3)    & 4.4 & 3.5 & 2308.9 & 0.4  & 143.9 & 2461.1 (0.38x)
         \\
 \Modelsp ($\text{L}_1$=6)   & 4.4 & 3.5 & 3463.4 & 0.4 & 143.9 & 3615.5 (0.56x)\\
 \Modelsp ($\text{L}_1$=9)    & 4.4 & 3.5 & 4617.9 & 0.4 & 143.9 & 4770.0 (0.73x) \\
  \Modelsp ($\text{L}_1$=12)    & 4.4 & 3.5 & 5772.3 & 0.4 & 143.9 & 5924.5 (0.91x) \\ 
   \hline
\end{tabular}
\caption{\#GFLOPs of FiD (base) and \Modelsp (base) over different stages in the model.}
\label{tab:flops-base}
\end{table*}

\begin{table*}[!tp]
\centering
\begin{tabular}{lcccccc}
\hline
Model & Retrieving & \multicolumn{1}{c}{\begin{tabular}[c]{@{}c@{}}Stage-1 \\ Reranking\end{tabular}} & \multicolumn{1}{c}{\begin{tabular}[c]{@{}c@{}}Reader \\ Encoding\end{tabular}} & \multicolumn{1}{c}{\begin{tabular}[c]{@{}c@{}}Stage-2 \\ Reranking\end{tabular}} & \multicolumn{1}{c}{\begin{tabular}[c]{@{}c@{}}Reader \\ Decoding\end{tabular}} & All        \\ \hline
FiD    & 4.4 & - & 17483.2 &  - & 2534.5 & 20022.0 (1.00x)  \\ \hline
  \Modelsp ($\text{L}_1$=6)    & 4.4 & 3.5 & 6993.3 & 0.6  & 510.0 & 7511.8 (0.38x)
         \\
 \Modelsp ($\text{L}_1$=12)   & 4.4 & 3.5 & 10489.9 & 0.6 & 510.0 & 11008.4 (0.55x)\\
 \Modelsp ($\text{L}_1$=18)    & 4.4 & 3.5 & 13986.5 & 0.6 & 510.0 & 14505.1 (0.72x) \\
  \Modelsp ($\text{L}_1$=24)    & 4.4 & 3.5 & 17483.2 & 0.6 & 510.0 & 18001.7 (0.90x) \\ 
   \hline
\end{tabular}
\caption{\#GFLOPs of FiD (large) and \Modelsp (large) over different stages in the model.}
\label{tab:flops-large}
\end{table*}

\subsection{FLOPs Computation}
\label{appendex:flops}

In this section we compute the FLOPs of each module\footnote{Our computation is based on https://github.com/google-research/electra/blob/master/flops\_computation.py}. The results are shown in Table \ref{tab:flops-base} and \ref{tab:flops-large} for base model and large model respectively. Before the computation, we first show some basic statistics on two benchmark datasets: the average question length is 20, and the average answer length is 5. For the reading part, the length of concatenated passage question pair is 250, number of input passages is $N_1=100$.

We first calculate the number of FLOPs of vanilla FiD model.
For the retrieving part, it contains both question encoding and passage similarity search. We only consider the former part as the latter part depends on the corpus size and search methods and is usually very efficient. The question encoding flops by BERT-based model is about 4.4 Gigaflops (GFLOPs). For the reading part, the encoding of each question passage pair takes about 57/174 GFLOPs for base/large model, and the encoding of 100 passages takes  5772/17483 GFLOPs. The decoder part only costs 714.2/2534.5 GFLOPs for base/large model since the average length of answer is very small. In summary, vanilla FiD base/large model costs 6491.0/20022.0 GFLOPs.  

For our model, the computation cost of retrieving part is the same as vanilla FiD. Since we set $N_0=1000$ and $N_1=100$, the GAT \citep{gat} computation in stage-1 reranking takes about 3.5 GFLOPs, and the stage-2 reranking takes only 0.4/0.6 GFLOPs for base/large model. For the reader encoding part, the computation cost depends on $L_1$ and $N_2$, which is analyzed in Section \ref{sec:analysis}. For the reader decoding part, where cross attention takes most of the computation, \Modelsp only takes about $N_2/N_1 = 1/5$ cost of vanilla FiD, which is 143.9/510.0 for base/large model respectively. The detailed flops are shown in Table \ref{tab:flops-base} and \ref{tab:flops-large}.

\end{document}